\documentclass{article}

    \usepackage[nonatbib, final]{tackling_climate_workshop_style}

\usepackage[utf8]{inputenc} 
\usepackage[T1]{fontenc}    
\usepackage{hyperref}       
\usepackage{url}            
\usepackage{booktabs}       
\usepackage{amsfonts}       
\usepackage{nicefrac}       
\usepackage{microtype}      
\usepackage{graphicx} 

\usepackage[style=numeric,sorting=none]{biblatex}
\bibliography{references}

\title{Multi-Source Temporal Attention Network for Precipitation Nowcasting}

\author{%
  Rafael Pablos Sarabia \\
  Aarhus University \\
  Cordulus \\
  \texttt{rpablos@cs.au.dk} \\
  \And
  Joachim Nyborg \\
  Cordulus \\
  \texttt{jn@cordulus.com} \\
  \And
  Morten Birk \\
  Cordulus \\
  \texttt{mb@cordulus.com} \\
  \AND
  Jeppe Liborius Sjørup \\
  Cordulus \\
  \texttt{jls@cordulus.com} \\
  \And
  Anders Lillevang Vesterholt \\
  Cordulus \\
  \texttt{alv@cordulus.com} \\
  \And
  Ira Assent \\
  Aarhus University \\
  \texttt{ira@cs.au.dk} \\
}

\begin{document}

\maketitle

\begin{abstract}
Precipitation nowcasting is crucial across various industries and plays a significant role in mitigating and adapting to climate change. We introduce an efficient deep learning model for precipitation nowcasting, capable of predicting rainfall up to 8 hours in advance with greater accuracy than existing operational physics-based and extrapolation-based models. Our model leverages multi-source meteorological data and physics-based forecasts to deliver high-resolution predictions in both time and space. It captures complex spatio-temporal dynamics through temporal attention networks and is optimized using data quality maps and dynamic thresholds. Experiments demonstrate that our model outperforms state-of-the-art, and highlight its potential for fast reliable responses to evolving weather conditions.
\end{abstract}

\section{Introduction}
Precipitation nowcasting can play a vital role in adapting to the impacts of climate change by providing accurate, high-resolution forecasts of rainfall intensity over a short period of up to 24 hours. This is a challenging task due to the sparse and non-Gaussian nature of precipitation. Additionally, climate change is making heavy precipitation events more frequent and altering their nature \cite{europe_precipitation}, increasing the uncertainty in predicting such rainfall. Since the 1870s, Denmark has experienced a 20\% increase in annual precipitation \cite{denmark_report}. More intense rainfall increases the risk of flooding, which can disrupt energy supplies by damaging infrastructure or prompting power outages for safety. Denmark's Climate Status and Outlook 2022 \cite{denmark_energy} notes that changes in precipitation, temperature, and wind have previously caused significant fluctuations in carbon dioxide equivalent (CO2e) emissions from electricity and heating sectors, varying by up to +/- 5 million tonnes of CO2e, mainly due to weather conditions like cold winters and fluctuating precipitation. This emphasizes the need for effective precipitation nowcasting to improve planning, prevention, and adaptation to the effects of climate change.

In agriculture, the main focus of Cordulus, precipitation nowcasting can contribute to combating climate change by reducing fuel consumption for unsuccessful trips to fields during unfavorable weather, optimizing timing of grain harvesting and drying to reduce energy use and dry matter loss, enhancing spray efficiency by dosage of products for weather conditions, and preventing product waste by scheduling plant protection treatments at optimal times.

Current operational methods for precipitation nowcasting include Numerical Weather Prediction (NWP) models and optical flow models like PySteps \cite{pysteps} and RainyMotion \cite{rainymotion}. NWP models solve mathematical equations \cite{nwp_guidelines} wrt. initial and boundary conditions. To improve forecast accuracy, ensemble NWP systems use multiple simulations with varying conditions \cite{ensemble_nwp}. Optical flow tracks radar echoes and projects their movement, assuming constant intensity. Both approaches have limitations. NWP models demand significant compute, especially for ensembles, which restricts their spatial and temporal resolution. Their long convergence time makes them ill-suited for short-term precipitation nowcasting, where accurate forecasts are needed for the initial hours. On the other hand, optical flow methods may overestimate precipitation and may not accurately cover all areas \cite{dgmr}.

Deep learning models demonstrate enhanced forecasting accuracy, particularly in per-grid-cell metrics, by optimizing directly with fewer biases. These models leverage advanced GPUs to produce forecasts within seconds \cite{metnet2} and excel at capturing complex, non-linear precipitation patterns due to their ability to analyze high-dimensional data. However, forecasting rain remains challenging due to the rapidly changing nature of atmospheric conditions and the variability in precipitation over short distances and times. Research has mainly focused on short-term forecasts (1 to 3 hours) using e.g. convolutional LSTMs \cite{metnet2}, spatio-temporal memory flows \cite{predrnn}, adversarial training \cite{dgmr}, latent diffusion models \cite{prediff, ldcast}, physical evolution schemes \cite{nowcastnet}, recurrent residual gates \cite{efsatrad}, and transformers \cite{dr2a-unet}.
In operational settings, MetNet-3 \cite{metnet3} and Pangu-Weather \cite{pangu-weather} are leading deep learning models in the US and Europe, respectively. MetNet-3 is a transformer-based model providing high-resolution precipitation forecasts for up to 24 hours. Pangu-Weather, also transformer-based with hierarchical temporal aggregation, offers forecasts of multiple variables for up to 168 hours but relies on ERA5 data, which has known biases for precipitation and much lower resolution \cite{graphcast}.

We introduce the first deep learning model for precipitation nowcasting for up to 8 hours that outperforms existing operational physics-based and extrapolation-based models in Denmark. Our model leverages multiple data sources of atmospheric conditions and physics-based forecasts, captures spatio-temporal dynamics, and is optimized via quality maps and dynamic thresholds.

\section{Approach: multi-source temporal attention}

The Danish Meteorological Institute (DMI) provides radar composite data that captures rainfall intensities at 10-minute intervals with a resolution of 500 meters per pixel \footnote{\url{https://opendatadocs.dmi.govcloud.dk/Data/Radar_Data}}. Still, its range is limited due to its ground-based nature (cf. Fig. \ref{quality-map}). To provide a more comprehensive view of the atmospheric state, we propose to complement the radar data with additional data from geostationary EUMETSAT satellites\footnote{European Organisation for the Exploitation of Meteorological Satellites  \url{https://user.eumetsat.int/data/satellites/meteosat-second-generation}} covering broader regions, but with lower resolution. We obtain physical properties from GFS satellite imagery\footnote{Global Forecast System managed by National Oceanic and Atmospheric Administration (NOAA), United States \url{https://www.emc.ncep.noaa.gov/emc/pages/numerical_forecast_systems/gfs.php}}, which provides forecasts with a spatial resolution of 0.25 degrees and hourly temporal resolution for the first five days of the forecast in addition to the current state of the atmosphere with derived physical measurements. We process this data, spanning from January 2022 to May 2024, to generate sequences of patches of size, resolution, and context optimized for each source to fit GPU memory, and use a sliding window with blackout periods to prevent data leakage (details in Appendix \ref{ap-data}).

\begin{figure}[h]
  \centering
  \includegraphics[width=0.75\linewidth]{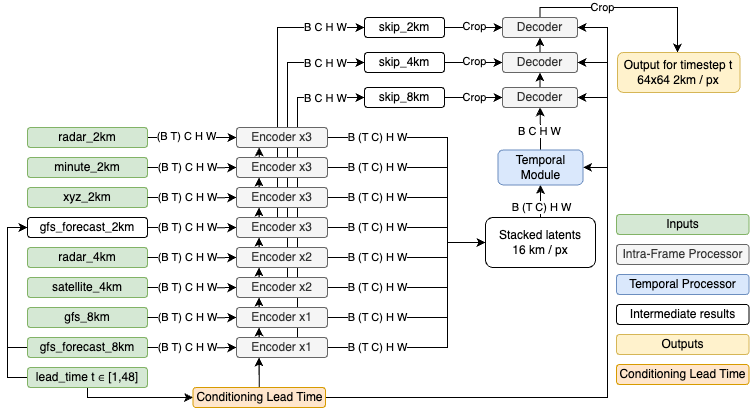}
  \caption{Proposed architecture, capable of simultaneously processing multiple data sources.}
  \label{tau-arch}
\end{figure}

\textbf{Architecture}
Recurrent networks suffer from poor computational efficiency, motivating us to leverage the Temporal Attention Unit (TAU) \cite{tau} which features a spatial encoder and a decoder for intra-frame features, with temporal modules stacked in between to extract time-dependent features. A residual connection between the encoder and decoder preserves spatial information. The temporal module, built for parallel processing, uses depth-wise convolutions, dilated depth-wise convolutions, and 1×1 convolutions to address long-range dependencies. Pooling across the spatial dimension and fully-connected layers across the temporal dimension allow to learn temporal variations.
The architecture in \cite{tau} has fixed number of timesteps and channels. To handle data sources with different timesteps, sizes, and resolutions, our encoder standardizes all inputs independently to the same resolution and size before feeding them into the temporal module. Additionally, our decoder includes a residual connection for each resolution and produces a single timestep for the specified lead time with channels to represent various rain intensities (Fig. \ref{tau-arch}). Instead of a continuous map, we predict probabilities in intensity bins to highlight both common light and rare heavy rainfall.

\textbf{Conditioning lead time}
To forecast sequences and prevent errors in intermediate forecasts from accumulating and affecting future predictions like in autoregressive approaches, we use conditioning lead time as in MetNet \cite{metnet}. With conditioning lead time, the model predicts a single lead time (specified as input) during each forward pass (cf. Fig. \ref{tau-arch}). Our model does not use predictions as inputs, allowing it to generate forecasts for the desired lead times independently and simultaneously.

\textbf{Loss function}
We employ cross-entropy loss for probabilistic forecasts to capture comprehensive information \cite{metnet, metnet2, metnet3}. We design classes of different rain intensities to have narrower ranges for lower precipitation values more frequently observed, while still encompassing rarer instances of heavy rain. Cross-entropy loss treats mispredictions equally, regardless of class. We instead propose weighing per-pixel loss by the difference between the indices of the target class and of the predicted class. 

\textbf{Quality map}
Averaging measured precipitation per pixel on radar maps from DMI reveals bias wrt. position of radar towers (detailed in Fig. \ref{quality-map}). We give greater weight to areas with reliable measurements by transforming these biases into a quality weight map. We still include lower-quality regions in the loss computation (with a lower weight) due to limited radar coverage available, unlike for MetNet \cite{metnet} and MetNet-2 \cite{metnet2}, where sufficient high-quality data is assumed.

\textbf{Dynamic thresholds}
Probabilistic outputs capture uncertainty, but some metrics and visualizations assume forecast intensities. An intensity value can be derived from the probability distribution over the rain classes as mean of the highest activated class. However, class activations are noisy, especially for highly unbalanced classes. 
To capture high precipitation events that are less likely and thus have lower predicted probability mass, we compute dynamic thresholds for each class and lead time after training, and consider a class activated when the predicted probability mass exceeds this threshold. Thresholds for rain intensities from probabilities are used in \cite{metnet2} but without specifying details.

\section{Experiments and results}
\label{sec:exp}
Our model has 15.2 million parameters, is trained for up to 50 epochs with 2,000 steps per epoch with random samples, using static learning rate of 3e-4, Adam optimizer with weight decay of 1e-3, and a batch size of 28 using PyTorch Lightning. The model with the lowest validation loss is selected. Training is conducted on an NVIDIA A10 Tensor Core 24 GB GPU and takes 24 hours to converge. We compare with state-of-the-art operational NWP models (Harmonie\cite{harmonie} and GFS) and an extrapolation-based method (PySTEPS). NWP forecasts are spatially and temporally interpolated to match the target resolution of 2km per pixel and 10-minute intervals. Assessment uses the Critical Success Index (CSI) metric at different thresholds, which primarily measures the accuracy of precipitation detection \cite{csi}, as commonly used in precipitation nowcasting \cite{metnet2, metnet3, prediff, dgmr}.

\begin{figure}[h]
  \centering
  \includegraphics[width=0.85\linewidth]{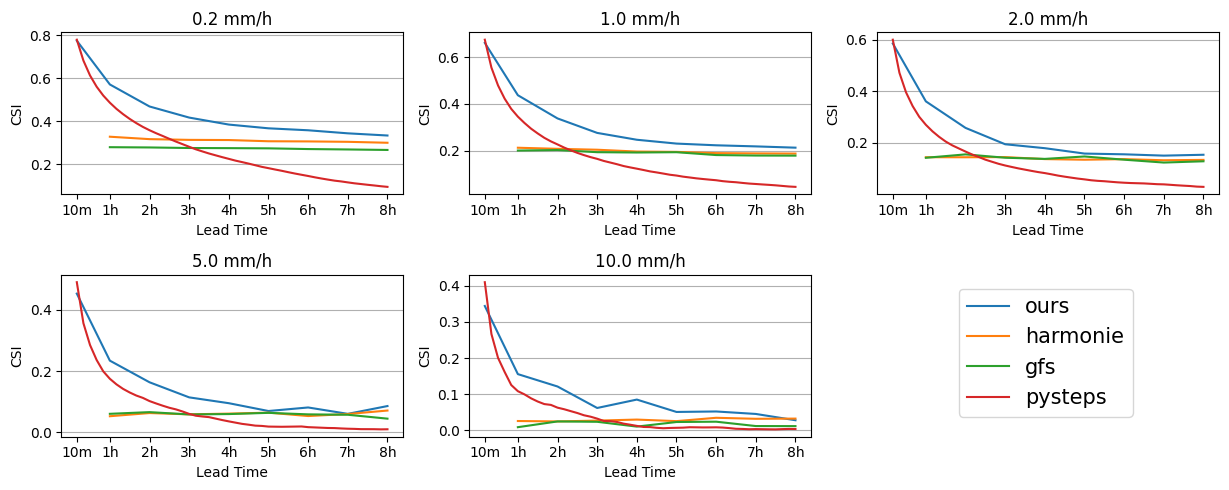}
  \caption{Critical Success Index (CSI) for various models across lead times.}
  \label{csi-test}
\end{figure}

Figure \ref{csi-test} shows results across rainfall intensities and lead times. Our model consistently outperforms extrapolation methods, which are constrained by their assumption of constant motion and intensity. Compared to NWP, our model exhibits a particularly large skill gap in the initial hours because of long NWP convergence time. Overall, our model achieves superior performance across lead times up to eight hours and for all thresholds. 
Our model uses as input NWP forecasts, specifically GFS forecasts, that are up to 3 hours old, but has higher temporal resolution of 10 minutes, compared to the hourly forecasts of NWP models, and can generate forecasts in minutes instead of in hours. 

\begin{figure}[h]
  \centering
  \includegraphics[width=0.77\linewidth]{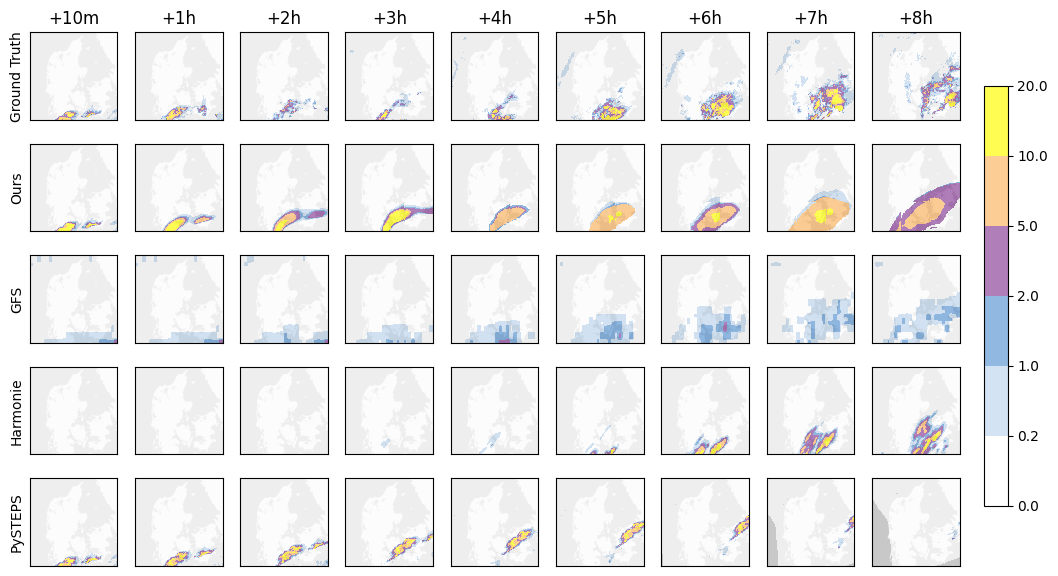}
  \caption{Sample ground truth, model prediction, GFS, Harmonie, and PySTEPS forecasts. Even though our model provides predictions at 10-minute intervals, hourly intervals are shown.}
  \label{viz-sample}
\end{figure}

A sample forecast, ground truth, GFS, Harmonie, and PySTEPS forecasts is shown in Figure \ref{viz-sample}. As lead time increases, uncertainty grows, resulting in more blurry precipitation patterns. Still, our model identifies the high precipitation forming in later lead times, unlike GFS. While our forecast appears more blurred than Harmonie, it actually improves accuracy by accounting for inherent uncertainty in predicting up to 8 hours ahead. Harmonie is noticeably shifted, which results in worse performance. PySTEPS accurately predicts the earlier lead times but quickly becomes ineffective due to the constant motion and intensity assumption.

\section{Discussion and conclusion}
We present a precipitation nowcasting model for Denmark that surpasses existing operational systems for up to 8 hours by leveraging multiple data sources, an advanced spatio-temporal architecture, optimized training with quality maps, and dynamic thresholds. Future work primarily involves expanding coverage to Europe radar from OPERA. Incorporating sparse observations from Cordulus' >4,000 European weather stations may correct radar observations with surface rain gauge measurements.

\begin{ack}
This work is partly funded by the Innovation Fund Denmark (IFD) under File No. 2052-00064B.
\end{ack}

\printbibliography

@article{dgmr,
	title = {Skilful precipitation nowcasting using deep generative models of radar},
	volume = {597},
	issn = {0028-0836, 1476-4687},
	url = {https://www.nature.com/articles/s41586-021-03854-z},
	doi = {10.1038/s41586-021-03854-z},
	language = {en},
	number = {7878},
	journal = {Nature},
	author = {Ravuri, Suman and Lenc, Karel and Willson, Matthew and Kangin, Dmitry and Lam, Remi and Mirowski, Piotr and Fitzsimons, Megan and Athanassiadou, Maria and Kashem, Sheleem and Madge, Sam and Prudden, Rachel and Mandhane, Amol and Clark, Aidan and Brock, Andrew and Simonyan, Karen and Hadsell, Raia and Robinson, Niall and Clancy, Ellen and Arribas, Alberto and Mohamed, Shakir},
	month = sep,
	year = {2021},
	pages = {672--677},
}

@ARTICLE{europe_precipitation,
       author = {{Nissen}, Katrin M. and {Ulbrich}, Uwe},
        title = "{Increasing frequencies and changing characteristics of heavy precipitation events threatening infrastructure in Europe under climate change}",
      journal = {Natural Hazards and Earth System Sciences},
         year = 2017,
        month = jul,
       volume = {17},
       number = {7},
        pages = {1177-1190},
          doi = {10.5194/nhess-17-1177-2017},
       adsurl = {https://ui.adsabs.harvard.edu/abs/2017NHESS..17.1177N}
}

@misc{denmark_report,
  author       = {{International Energy Agency}},
  title        = {{Denmark: Climate Resilience Policy Indicator}},
  year         = 2022,
  url          = {https://www.iea.org/reports/denmark-climate-resilience-policy-indicator},
  publisher    = {{International Energy Agency}},
  license      = {CC BY 4.0}
}

@misc{denmark_energy,
  author       = {{Danish Energy Agency}},
  title        = {{Denmark's Climate Status and Outlook 2022}},
  year         = 2022,
  url          = {https://ens.dk/sites/ens.dk/files/Forskning_og_udvikling/cso22_-_english_translation_of_kf22_hovedrapport.pdf},
  publisher    = {{Danish Energy Agency}},
  address      = {Copenhagen}
}

@article{rainymotion,
	title = {Optical flow models as an open benchmark for radar-based precipitation nowcasting (rainymotion v0.1)},
	volume = {12},
	issn = {1991-9603},
	url = {https://gmd.copernicus.org/articles/12/1387/2019/},
	doi = {10.5194/gmd-12-1387-2019},
	language = {en},
	number = {4},
	journal = {Geoscientific Model Development},
	author = {Ayzel, Georgy and Heistermann, Maik and Winterrath, Tanja},
	month = apr,
	year = {2019},
	pages = {1387--1402},
}

@misc{metnet,
	title = {{MetNet}: {A} {Neural} {Weather} {Model} for {Precipitation} {Forecasting}},
	shorttitle = {{MetNet}},
	url = {http://arxiv.org/abs/2003.12140},
	language = {en},
	publisher = {arXiv},
	author = {Sønderby, Casper Kaae and Espeholt, Lasse and Heek, Jonathan and Dehghani, Mostafa and Oliver, Avital and Salimans, Tim and Agrawal, Shreya and Hickey, Jason and Kalchbrenner, Nal},
	month = mar,
	year = {2020},
	note = {arXiv:2003.12140 [physics, stat]},
}

@misc{metnet2,
	title = {Skillful {Twelve} {Hour} {Precipitation} {Forecasts} using {Large} {Context} {Neural} {Networks}},
	url = {http://arxiv.org/abs/2111.07470},
	language = {en},
	publisher = {arXiv},
	author = {Espeholt, Lasse and Agrawal, Shreya and Sønderby, Casper and Kumar, Manoj and Heek, Jonathan and Bromberg, Carla and Gazen, Cenk and Hickey, Jason and Bell, Aaron and Kalchbrenner, Nal},
	month = nov,
	year = {2021},
	note = {arXiv:2111.07470 [physics]},
}

@misc{metnet3,
	title = {Deep {Learning} for {Day} {Forecasts} from {Sparse} {Observations}},
	url = {http://arxiv.org/abs/2306.06079},
	language = {en},
	publisher = {arXiv},
	author = {Andrychowicz, Marcin and Espeholt, Lasse and Li, Di and Merchant, Samier and Merose, Alexander and Zyda, Fred and Agrawal, Shreya and Kalchbrenner, Nal},
	month = jul,
	year = {2023},
	note = {arXiv:2306.06079 [physics]},
}

@misc{pangu-weather,
	title = {Pangu-{Weather}: {A} {3D} {High}-{Resolution} {Model} for {Fast} and {Accurate} {Global} {Weather} {Forecast}},
	shorttitle = {Pangu-{Weather}},
	url = {http://arxiv.org/abs/2211.02556},
	language = {en},
	publisher = {arXiv},
	author = {Bi, Kaifeng and Xie, Lingxi and Zhang, Hengheng and Chen, Xin and Gu, Xiaotao and Tian, Qi},
	month = nov,
	year = {2022},
	note = {arXiv:2211.02556 [physics]},
}

@misc{graphcast,
	title = {{GraphCast}: Learning skillful medium-range global weather forecasting},
	url = {http://arxiv.org/abs/2212.12794},
	shorttitle = {{GraphCast}},
	number = {{arXiv}:2212.12794},
	publisher = {{arXiv}},
	author = {Lam, Remi and Sanchez-Gonzalez, Alvaro and Willson, Matthew and Wirnsberger, Peter and Fortunato, Meire and Alet, Ferran and Ravuri, Suman and Ewalds, Timo and Eaton-Rosen, Zach and Hu, Weihua and Merose, Alexander and Hoyer, Stephan and Holland, George and Vinyals, Oriol and Stott, Jacklynn and Pritzel, Alexander and Mohamed, Shakir and Battaglia, Peter},
	date = {2023-08-04},
	eprinttype = {arxiv},
	eprint = {2212.12794 [physics]},
}

@misc{nwp_guidelines,
	title = {Guidelines for {Ensemble} {Prediction} {System}},
	publisher = {World Meteorological Organization},
	author = {Mylne, K and Chen, J and Erfani, A and Hamill, T and Richardson, D and Vannitsem, Stephane and Wang, Y and Zhu, Y},
	year = {2022},
}

@article{ensemble_nwp,
	title = {Ensemble {Forecasting} at {NCEP} and the {Breeding} {Method}},
	volume = {125},
	issn = {0027-0644, 1520-0493},
	url = {http://journals.ametsoc.org/doi/10.1175/1520-0493(1997)125<3297:EFANAT>2.0.CO;2},
	doi = {10.1175/1520-0493(1997)125<3297:EFANAT>2.0.CO;2},
	language = {en},
	number = {12},
	journal = {Monthly Weather Review},
	author = {Toth, Zoltan and Kalnay, Eugenia},
	month = dec,
	year = {1997},
	pages = {3297--3319},
}

@article{pysteps,
	title = {Pysteps: an open-source {Python} library for probabilistic precipitation nowcasting (v1.0)},
	volume = {12},
	issn = {1991-9603},
	shorttitle = {Pysteps},
	url = {https://gmd.copernicus.org/articles/12/4185/2019/},
	doi = {10.5194/gmd-12-4185-2019},
	language = {en},
	number = {10},
	journal = {Geoscientific Model Development},
	author = {Pulkkinen, Seppo and Nerini, Daniele and Pérez Hortal, Andrés A. and Velasco-Forero, Carlos and Seed, Alan and Germann, Urs and Foresti, Loris},
	month = oct,
	year = {2019},
	pages = {4185--4219},
}

@misc{predrnn,
	title = {{PredRNN}: A Recurrent Neural Network for Spatiotemporal Predictive Learning},
	url = {http://arxiv.org/abs/2103.09504},
	shorttitle = {{PredRNN}},
	number = {{arXiv}:2103.09504},
	publisher = {{arXiv}},
	author = {Wang, Yunbo and Wu, Haixu and Zhang, Jianjin and Gao, Zhifeng and Wang, Jianmin and Yu, Philip S. and Long, Mingsheng},
	date = {2022-04-09},
	eprinttype = {arxiv},
	eprint = {2103.09504 [cs]},
}

@misc{prediff,
      title={PreDiff: Precipitation Nowcasting with Latent Diffusion Models}, 
      author={Zhihan Gao and Xingjian Shi and Boran Han and Hao Wang and Xiaoyong Jin and Danielle Maddix and Yi Zhu and Mu Li and Yuyang Wang},
      year={2023},
      eprint={2307.10422},
      archivePrefix={arXiv},
      primaryClass={cs.LG}
}

@misc{ldcast,
	title = {Latent diffusion models for generative precipitation nowcasting with accurate uncertainty quantification},
	url = {http://arxiv.org/abs/2304.12891},
	number = {{arXiv}:2304.12891},
	publisher = {{arXiv}},
	author = {Leinonen, Jussi and Hamann, Ulrich and Nerini, Daniele and Germann, Urs and Franch, Gabriele},
	date = {2023-04-25},
	eprinttype = {arxiv},
	eprint = {2304.12891 [physics]},
}

@article{nowcastnet,
	title = {Skilful nowcasting of extreme precipitation with {NowcastNet}},
	volume = {619},
	issn = {0028-0836, 1476-4687},
	url = {https://www.nature.com/articles/s41586-023-06184-4},
	doi = {10.1038/s41586-023-06184-4},
	language = {en},
	number = {7970},
	journal = {Nature},
	author = {Zhang, Yuchen and Long, Mingsheng and Chen, Kaiyuan and Xing, Lanxiang and Jin, Ronghua and Jordan, Michael I. and Wang, Jianmin},
	month = jul,
	year = {2023},
	pages = {526--532},
}

@misc{efsatrad,
	title = {Transformer-based nowcasting of radar composites from satellite images for severe weather},
	url = {http://arxiv.org/abs/2310.19515},
	number = {{arXiv}:2310.19515},
	publisher = {{arXiv}},
	author = {Küçük, Çağlar and Giannakos, Apostolos and Schneider, Stefan and Jann, Alexander},
	date = {2023-10-30},
	eprinttype = {arxiv},
	eprint = {2310.19515 [physics]},
}

@article{dr2a-unet,
	title = {Radar-Based Precipitation Nowcasting Based on Improved U-Net Model},
	volume = {16},
	issn = {2072-4292},
	url = {https://www.mdpi.com/2072-4292/16/10/1681},
	doi = {10.3390/rs16101681},
	number = {10},
	journaltitle = {Remote Sensing},
	author = {Tan, Youwei and Zhang, Ting and Li, Leijing and Li, Jianzhu},
	date = {2024},
}

@misc{tau,
	title = {Temporal Attention Unit: Towards Efficient Spatiotemporal Predictive Learning},
	url = {http://arxiv.org/abs/2206.12126},
	shorttitle = {Temporal Attention Unit},
	number = {{arXiv}:2206.12126},
	publisher = {{arXiv}},
	author = {Tan, Cheng and Gao, Zhangyang and Wu, Lirong and Xu, Yongjie and Xia, Jun and Li, Siyuan and Li, Stan Z.},
	date = {2023-04-12},
	eprinttype = {arxiv},
	eprint = {2206.12126 [cs]},
}

@article{csi,
	title = {The Critical Success Index as an Indicator of Warning Skill},
	volume = {5},
	url = {https://journals.ametsoc.org/view/journals/wefo/5/4/1520-0434_1990_005_0570_tcsiaa_2_0_co_2.xml},
	doi = {10.1175/1520-0434(1990)005<0570:TCSIAA>2.0.CO;2},
	pages = {570 -- 575},
	number = {4},
	journaltitle = {Weather and Forecasting},
	author = {Schaefer, Joseph T.},
	date = {1990},
	note = {Place: Boston {MA}, {USA}
Publisher: American Meteorological Society},
}

@article{harmonie,
	title = {The {HARMONIE}–{AROME} Model Configuration in the {ALADIN}–{HIRLAM} {NWP} System},
	volume = {145},
	url = {https://journals.ametsoc.org/view/journals/mwre/145/5/mwr-d-16-0417.1.xml},
	doi = {10.1175/MWR-D-16-0417.1},
	pages = {1919 -- 1935},
	number = {5},
	journaltitle = {Monthly Weather Review},
	author = {Bengtsson, Lisa and Andrae, Ulf and Aspelien, Trygve and Batrak, Yurii and Calvo, Javier and Rooy, Wim de and Gleeson, Emily and Hansen-Sass, Bent and Homleid, Mariken and Hortal, Mariano and Ivarsson, Karl-Ivar and Lenderink, Geert and Niemelä, Sami and Nielsen, Kristian Pagh and Onvlee, Jeanette and Rontu, Laura and Samuelsson, Patrick and Muñoz, Daniel Santos and Subias, Alvaro and Tijm, Sander and Toll, Velle and Yang, Xiaohua and Køltzow, Morten Ødegaard},
	date = {2017},
	note = {Place: Boston {MA}, {USA}
Publisher: American Meteorological Society},
}

\newpage
\appendix
\counterwithin{figure}{section}
\counterwithin{table}{section}

\section{Quality map}

\begin{figure}[h]
  \centering
  \includegraphics[width=0.35\linewidth]{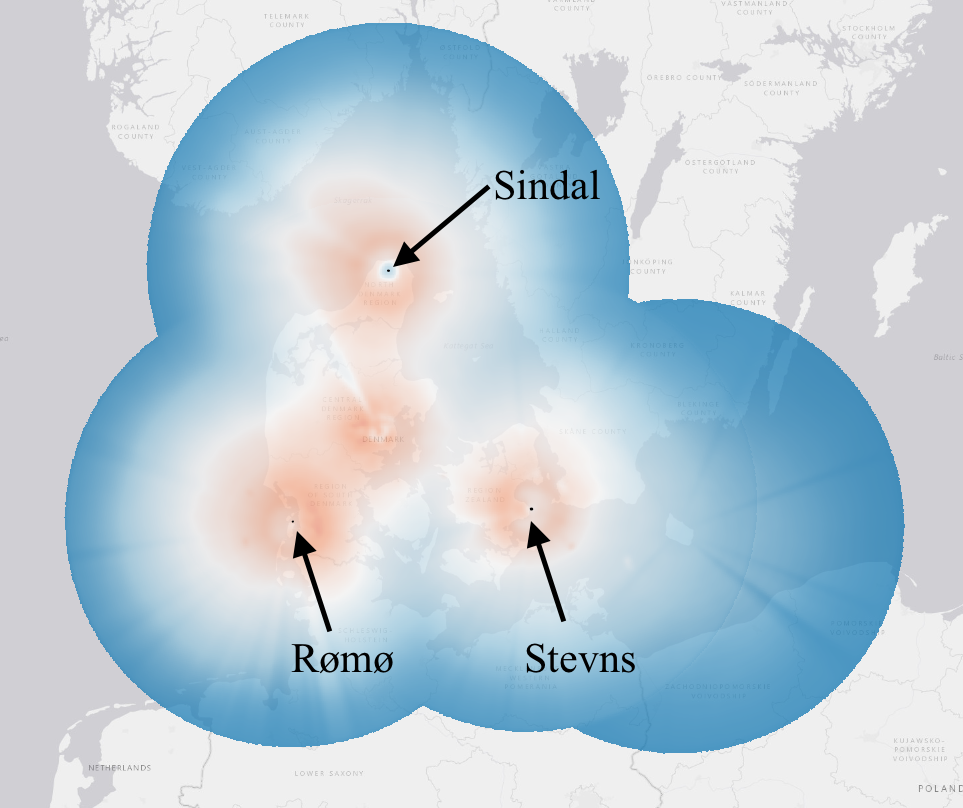}
  \caption{Estimated quality map over Denmark, created by averaging measurements per pixel over a 3-year period. The locations of the three active radar stations, as indicated by DMI, are also shown.}
  \label{quality-map}
\end{figure}

Figure \ref{quality-map} shows that regions near the radar towers and those farthest from them tend to under-represent rainfall. Additionally, it highlights some bias in the central region of Denmark due to a radar station in Virring. 

\section{Data preparation}
\label{ap-data}

\paragraph{Inputs and outputs}
All data sources have been processed to generate sequences of patches to train and evaluate the models in the GPU. The labels for the model are future radar maps covering a region of $128 \times 128 km^2$ and the final forecast for Denmark is constructed by combining individual model predictions obtained with each forward pass for a designated region. The input dimensions are therefore determined based on the target dimensions and the additional context needed (60-80 km per hour of forecast if available \cite{metnet2}). For the Global Forecast System (GFS) data, 122 channels are selected, representing various measurements and pressure levels to capture the current atmospheric state. Additionally, the GFS hourly precipitation forecast at a much lower spatial resolution is also used as input. It is important to note that these forecasts are derived from physical models, and in addition to their lower spatial and temporal resolutions, they also carry uncertainties and limited accuracy. Overall, these inputs aim to replicate those used in the MetNet models \cite{metnet, metnet2, metnet3}, utilizing the data available for Denmark. Tables \ref{output-table} and \ref{input-table} specify the output for the model and the corresponding inputs from the different data sources.

\begin{table}[h]
\caption{Outputs for the precipitation nowcasting model}
  \label{output-table}
  \centering
  \begin{tabular}{llllllll}
    \toprule
Variable & Source & \begin{tabular}[c]{@{}l@{}}Size\\ (px)\end{tabular} & \begin{tabular}[c]{@{}l@{}}Res. \\ (km/px)\end{tabular} & \begin{tabular}[c]{@{}l@{}}Context\\ (km)\end{tabular} & \begin{tabular}[c]{@{}l@{}}Timesteps\\ (min)\end{tabular} & Channels \\
    \midrule
target\_2km & DMI & 64 & 2 & N/A & {[}10,20,...,480{]} & 1 \\
\bottomrule
\end{tabular}
\end{table}

\begin{table}[h]
\caption{Inputs for the precipitation nowcasting model}
  \label{input-table}
  \centering
  \begin{tabular}{llllllll}
    \toprule
Variable & Source & \begin{tabular}[c]{@{}l@{}}Size\\ (px)\end{tabular} & \begin{tabular}[c]{@{}l@{}}Res. \\ (km/px)\end{tabular} & \begin{tabular}[c]{@{}l@{}}Context\\ (km)\end{tabular} & \begin{tabular}[c]{@{}l@{}}Timesteps\\ (min)\end{tabular} & Channels \\
    \midrule
radar\_2km & DMI & 288 & 2 & 112 & {[}-90,-80,...,0{]} & 1 \\
radar\_4km & DMI & 288 & 4 & 512 & {[}0{]} & 1 \\
satellite\_4km & EUMETSAT & 288 & 4 & 512 & {[}-30,-15,0{]} & 11 \\
gfs\_8km & GFS & 144 & 8 & 512 & {[}0{]} & 122 \\
gfs\_forecast\_8km & GFS & 144 & 8 & 512 & {[}60,120,...,480{]} & 1 \\
xyz\_2km & - & 288 & 2 & 112 & N/A & 3 \\
minute\_2km & - & 288 & 2 & 112 & N/A & 1 \\
\bottomrule
\end{tabular}
\end{table}

\paragraph{Splits}
The data available for training and evaluating the model spans from January 2022 to May 2024. The training, validation, and testing data are obtained based on cycles of multiple hours, where each split is assigned every 200 hours, with a 12-hour blackout period between them to prevent data leakage. Since baseline forecasts are only available from May 2023 onward, data collected before this period is exclusively used for training. 

Samples are created with a sliding window technique applied over the temporal dimension for a smaller patch or region, ensuring that the corresponding spatial dimensions match the model's specifications. The training dataset is limited to 1 million sequences or samples, which are selected using importance sampling to prioritize instances with rainfall to improve training efficiency \cite{dgmr}. The validation set contains 5,000 randomly selected samples and the test set includes all 13,188 samples.

\end{document}